%% file: root.tex
\documentclass[letterpaper, 10 pt, conference]{ieeeconf} 

\IEEEoverridecommandlockouts                              

\usepackage{graphics} 
\usepackage{epsfig} 
\usepackage{times} 
\usepackage{amsmath} 
\usepackage{amssymb}  
\usepackage{biblatex}
\usepackage{multirow}
\usepackage{multicol}
\usepackage{hyperref}
\usepackage{booktabs}
\usepackage{enumerate}
\usepackage{color}
\usepackage{tabularx}
\usepackage{wrapfig}
\usepackage{hyperref}
\usepackage{pifont}
\usepackage{algpseudocode}
\usepackage{algorithm}
\usepackage{amsmath}
\usepackage{adjustbox}
\usepackage[usenames,dvipsnames,table]{xcolor}
\usepackage[font=small,labelfont=bf]{caption}

\usepackage{subcaption} 

\definecolor{darkgrey}{rgb}{0.5,0.5,0.5}

\input{text/macros}

\title{
\LARGE\bf Scaling Cross-Embodiment World Models for Dexterous Manipulation
}

\author{
Zihao He$^{1,2*}$, Bo Ai$^{1,3*}$, Tongzhou Mu$^1$, Yulin Liu$^1$, Weikang Wan$^1$, Jiawei Fu$^1$,  \\
Yilun Du$^4$, Henrik I. Christensen$^1$, and Hao Su$^{5}$ \vspace{3pt}
\\  
$^1$UC San Diego \quad 
$^2$Shanghai Jiao Tong University \quad 
$^3$Stanford University \quad  \\
$^4$Harvard University \quad 
$^5$Sudo AI GmbH \quad
$^*$Equal contribution \\ 
\vspace{3pt}
\authorblockA{\textbf{\textcolor{magenta}{\url{https://alan-heoooh.github.io/dexwm.html}}}}
\vspace{-4pt}
}

\begin{document}

\maketitle
\thispagestyle{empty}
\pagestyle{empty}

\setlength{\abovedisplayskip}{1pt}
\setlength{\belowdisplayskip}{1pt}

\begin{abstract}

Cross-embodiment learning seeks to build generalist robots that learn from and operate across diverse morphologies, but differences in kinematics and action spaces hinder data sharing and control transfer. We ask: What structure can be shared across embodiments despite these differences? We argue that the physical interactions they induce can be modeled in a shared geometric space, allowing world models to provide a common interface for learning and control. To realize this idea, we represent human and robot hands as sets of 3D particles and define actions as end-effector particle displacement fields. This representation abstracts away embodiment-specific joint spaces while preserving the geometry and motion relevant to physical interaction. We train a graph-based world model on random interaction data from diverse simulated robot hands and real human hands, and integrate it with model-predictive control for deployment on new hardware. Experiments on rigid and deformable manipulation reveal three findings: increasing the diversity of training embodiments improves generalization to unseen hands; appropriately combining simulated and real-world data outperforms either source alone; and the same learned model enables effective control on robotic hands with distinct kinematics and degrees of freedom. These results position particle-based world models as a shared interface for learning from and for heterogeneous embodiments.


\end{abstract}


\input{text/1_introduction}
\input{text/2_related_works}

\input{text/3_method}

\input{text/4_experiments}

\input{text/6_conclusions}
\printbibliography

\end{document}

%% file: text/macros.tex
\newcommand{\secref}[1]{Section~\ref{#1}}
\renewcommand{\eqref}[1]{Eqn~\ref{#1}}
\newcommand{\figref}[1]{Figure~\ref{#1}}

\newcommand{\tabref}[1]{Table~\ref{#1}}

\newcommand{\eqnref}[1]{\eqref{#1}}

\newcommand{\ie}{\textrm{i.e.}}
\newcommand{\eg}{\textrm{e.g.}}

\def\T{\mathcal{T}}
\def\T{\mathcal{J}}
\def\T{\mathcal{T}}

\def\weblink{\urllink[pre = \bgroup\bf, post = \egroup]}

  {\begin{mdframed}[topline=false, bottomline=false, rightline=false,%
    linewidth=1pt,linecolor=black,%
    leftmargin=1cm,rightmargin=1cm,%
    innerleftmargin=10pt,innerrightmargin=10pt,%
    innertopmargin=0pt,innerbottommargin=0pt,%
    skipabove=\baselineskip,skipbelow=\baselineskip]%
   \fontfamily{ppl}\selectfont
  }%
  {\end{mdframed}}

\def\pushingtask{\textit{Object Pushing}}
\def\reshapingtask{\textit{Plasticine Reshaping}}

\def\qone{\textbf{Q1}}
\def\qtwo{\textbf{Q2}}
\def\qthree{\textbf{Q3}}

\def\fingerspinch{\texttt{FingersPinch}}
\def\palmpress{\texttt{PalmPress}}
\def\thumbpinch{\texttt{ThumbPinch}}

%% file: text/1_introduction.tex
\section{Introduction}\label{sec:introduction}


Cross-embodiment learning seeks to build generalist robots that learn from and operate across diverse physical embodiments. Yet every embodiment expresses action through a different kinematic structure and control space, fragmenting interaction data and preventing direct transfer. This challenge is becoming increasingly important as large-scale robot deployments produce heterogeneous datasets across robot platforms and hardware generations~\cite{openx, rh20t}. Prior progress has shown embodiment-level generalization in locomotion~\cite{ai2025towards} and in manipulation with parallel grippers~\cite{Black2024pizero, Kim2024openvla, cage, octo_2023, grover2025enhancing, intelligence2026pi}, whereas dexterous manipulation has largely been limited to grasping \cite{dro, fang2025anydexgrasp} and in-hand reorientation~\cite{getzero}. Extending cross-embodiment learning to non-prehensile and deformable-object manipulation remains challenging because these tasks require both reasoning about complex object dynamics and fine-grained contact control.

Dexterous hands offer a compelling case for cross-embodiment learning. Despite the challenge of contact-rich, high-DoF control, multifingered robot hands are morphologically similar to one another and to human hands. This anthropomorphism suggests that cross-embodiment datasets, including not only robot datasets but also human–object interactions, can be mutually informative. This raises two central questions: what \textit{invariant knowledge} about the external world and contact interactions underlies purposeful action across distinct kinematics and control spaces, and how should state and action be \textit{represented} so that human and robot data become jointly useful? Our key insight is that joint-space actions are embodiment-specific, but the physical interactions they induce can be modeled in a shared geometric space. We argue that world models~\cite{ai2025review, tenenbaum2011howto}, which predict how physical states evolve under action, can provide a common interface for learning and control in this space.
The central challenge is therefore to design shared state and action representations that abstract away embodiment-specific details while preserving the geometry and motion that govern physical interaction, allowing a single predictive model to learn from heterogeneous data and enable model-based control across diverse kinematic structures.

To this end, we represent both human and robot hands as \emph{particles} (\ie, 3D point sets), with actions defined as particle displacements. A graph-based dynamics model~\cite{robocook, robopack, robocraft, shi2023robocraft, zhang2024adaptigraph} predicts particle motion while exploiting spatial locality and equivariance. We co-train the model on simulated robot–object interaction data and real human–object interaction data and study how generalization of the learned model scales with the morphological diversity in the training domain. For control, we sample robot joint actions, which are then converted to particle action representation via forward kinematics, to enable model-predictive control in the particle space. 
This action abstraction unifies control problems across embodiments, allowing the learned model to be deployed on hands with varied control spaces without motion retargeting or expert demonstration collection. 

We evaluate learned world models both in simulation and on real hardware. In simulation, we observe an \emph{embodiment scaling} trend~\cite{ai2025towards}: training on more simulated hands consistently improves generalization to unseen embodiments. In the real world, we find that models trained solely on human data can already transfer to robotic hands despite the embodiment gap, and that incorporating an appropriate amount of simulation data further improves both predictive accuracy and planning performance. Our best model, co-trained on simulated robot data and real human data, enables a 6-DoF PSYONIC Ability Hand and a 12-DoF Robot Era XHand to successfully perform deformable object manipulation. These results demonstrate the promise of world models as a unifying interface for cross-embodiment learning and generalization in robot manipulation.






%% file: text/2_related_works.tex
\section{Related Works}

\begin{figure*}[t]
    \centering
    \includegraphics[width=0.85\textwidth]{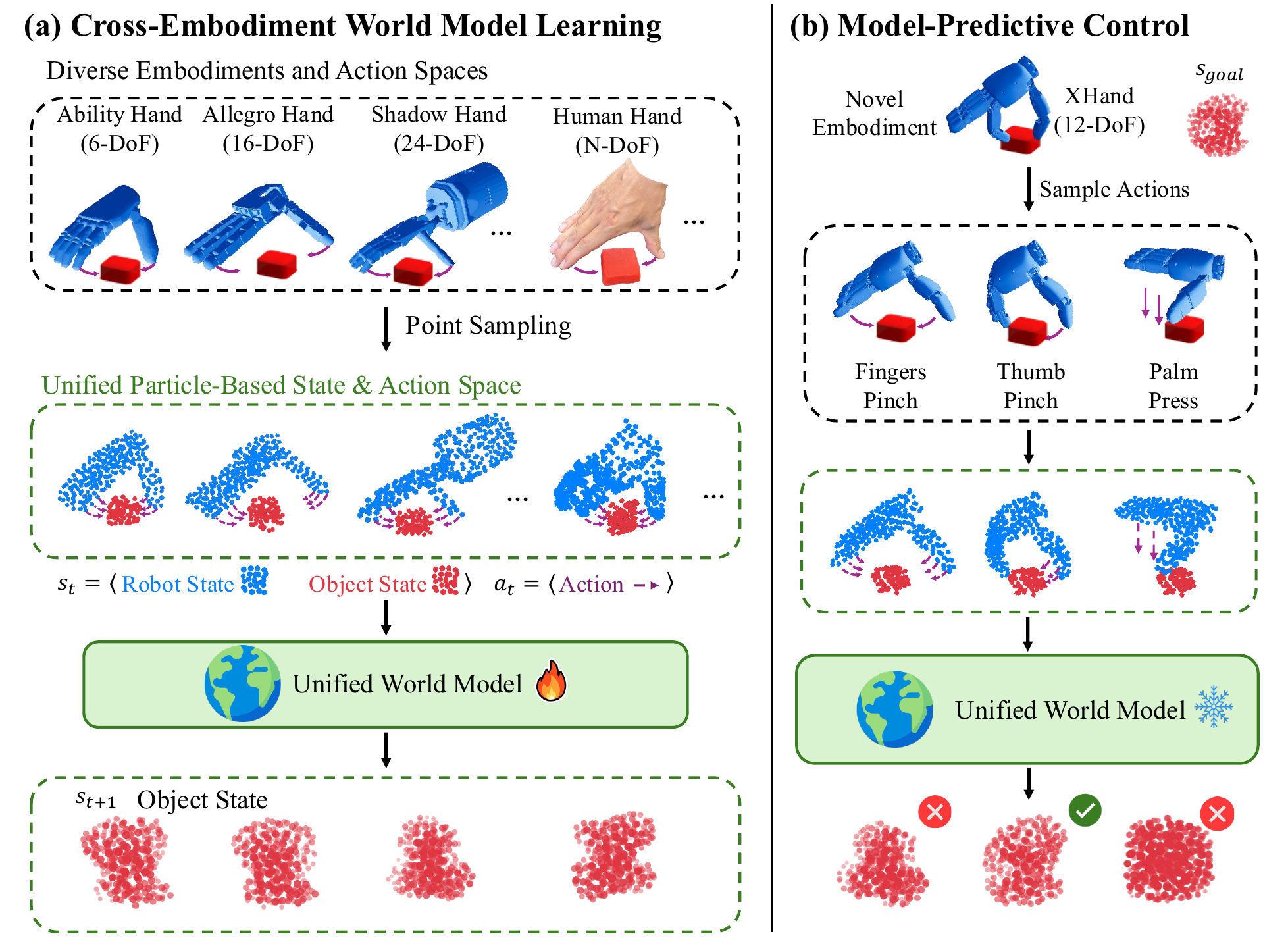} 
    \caption{\textbf{Overall framework}.  Our key idea is to represent both embodiments and objects as 3D particles, and actions as end-effector particle displacement fields. These state–action abstractions unify data and control across embodiments. (a) We train world models on random interaction data from diverse robot hands in simulation and from human demonstrations in the real world. (b) At deployment, joint action samples are mapped into displacement fields via forward kinematics, rolled out by the world model for prediction, and the optimal trajectory is executed on the target hardware. We illustrate a planning horizon of 1 here for simplicity.}
    \label{fig:framework} \vspace{-15pt}
\end{figure*}

\subsection{Cross-Embodiment Learning}



One key goal of cross-embodiment learning is to learn from diverse embodiments, with learning from human data being a special case. Most existing approaches are model-free, which learn mappings from observations to actions end-to-end via reinforcement learning (RL) or behavior cloning (BC). Prior work includes using human demonstrations to guide RL for dexterous manipulation~\cite{dexmv, chen2024vividex}, cross-embodiment RL training for locomotion~\cite{bohlinger2024onepolicy}, and scaling to many embodiments through a combination of RL and BC~\cite{ai2025towards, getzero}. When high-quality demonstrations are available, BC can provide stable learning signals~\cite{Punamiya2026egoverse, kareer2025emergence}.
Human demonstrations can also be adapted to robots via motion retargeting for anthropomorphic robots~\cite{qiu2025-humanpolicy}.  These works either require extensive RL training in simulation, limited to domains where the sim-to-real gap is moderate, or require expert demonstrations in the real world. 

Model-based approaches instead learn world models that explicitly predict action outcomes and have demonstrated strong performance on tasks requiring fine-grained control~\cite{ai2025review}. Prior work learns particle-based dynamics from human–object interactions but uses the model only to guide RL exploration~\cite{hong2025learning}, while concurrent PointWorld represents parallel-jaw gripper actions as 3D point flows and learns scene dynamics from large-scale robotic manipulation data~\cite{huang2026pointworld}. In contrast, we study world-model scaling across articulated human and robotic hands, jointly leveraging simulated robot interactions and real human data to enable direct model-predictive control of hands with distinct kinematics and degrees of freedom for dexterous manipulation tasks.

\subsection{World Model Learning for Robotic Manipulation}

World models, predictive models that forecast the effects of actions, are central to model-based robotic control \cite{ai2025review}. Their effectiveness depends heavily on state representation. Pixel-based models~\cite{ebert2018visual, yang2023learning, du2023learning} can exploit large-scale visual data but struggle to produce physically accurate predictions in contact-rich settings with extensive training. Particle-based models are more physically grounded and incorporate stronger inductive biases. They have enabled manipulation of clothes~\cite{Lin2021learningvisible, tian2025uniclothdiff}, ropes~\cite{zhang2024dynamics}, granular object piles \cite{wang2023dynamic}, and plasticine~\cite{robocraft, robocook}, but are often trained on single-embodiment data from parallel jaw grippers. This work seeks to establish particle-based world models for complex dexterous manipulation and position them as a paradigm for cross-embodiment learning.



\subsection{Dexterous Manipulation}

Dexterous manipulation is a long-standing challenge in robotics~\cite{salisbury1982articulated,mordatch2012contact}, largely due to the high degrees of freedom of multi-fingered hands and complex contact patterns. 
Classic control methods rely on analytical object models~\cite{rus1999hand,suh2025dexterous, Pang2023global}, which may not capture hard-to-model factors, such as frictional contact or actuator drift, and can be hard to obtain for deformable objects. 
Recent learning-based approaches, including RL and BC, have shown success in rigid object manipulation, such as grasping~\cite{wan2023unidexgrasp++,fang2025anydexgrasp}, in-hand reorientation~\cite{handa2023dextreme, From-Simple-to-Complex-Skills}, and tool use~\cite{shaw2024bimanual, wan2025lodestar, feng2025dqrise}. 
However, manipulating deformable objects using multi-fingered hands remains under-explored~\cite{dexdeform,zhaole2024dexdlo}, due to the combined challenges of high-dimensional control and complex object dynamics. 
In this work, we learn a model of environment dynamics and integrate it with model-predictive control. By introducing embodiment-agnostic state and action representations, we enable learning from both robotic and human data, allowing a single model to operate over diverse dexterous hands.

%% file: text/3_method.tex
\section{Method}

Our goal is to enable dexterous manipulation skills from and for diverse robotic hands. 
We formalize the general problem as follows. At each time step \(t\), the end effector is in configuration \(q_t \in \mathbb{R}^{n_e}\), where \(n_e\) is the number of degrees of freedom of embodiment \(e\), and the object is in state $s_{obj}$. The world state includes the state of both the robot and the object, $s_t = \langle q_t, s_{obj} \rangle$. The robot takes an action $u_t$, and the world transits to a new state $s_{t+1}$. 
The objective is to find an action sequence of length $H$, $u_{0:H-1}$, that minimizes a cost function $\mathcal{J}$: 
\begin{equation}
u_{0:H-1}^* = \arg\min_{u_{0:H-1} \in \mathcal{U}} 
\; \mathcal{J}\big(\mathcal{T}(s_0, u_{0:H-1}), s_g \big),
\label{eq:opt_compact}
\end{equation}
where $\mathcal{T}(s_0, u_{0:H-1})$ is the state reached after applying the sequence to the dynamics, 
and $s_g$ is the target state.

What is the shared underlying process across different embodiments for these control problems? 
Our key insight is that the underlying physical interaction process, captured by $\mathcal{T}$, is universal. 
However, approximating $\mathcal{T}$ is challenging due to the varying dimensions of the robot configuration 
$q_t \in \mathbb{R}^{n_e}$ and action $u_t \in \mathbb{R}^{n_e}$, which depend on the embodiment $e$, as well 
as the differences in kinematic and geometric structures that shape the environment dynamics. 
Therefore, we aim to unify state and action representations to learn embodiment-agnostic world models, which hold the potential to scale with cross-embodiment datasets. 


We next discuss the high-level framework of cross-embodiment model learning and planning (\secref{sec:world-model}), state estimation (\secref{sec:perception}), world model architecture (\secref{sec:world-model-details}), and model-based control (\secref{sec:control}).

\subsection{Cross-Embodiment World Model Learning and Planning} \label{sec:world-model}

We define a \textbf{particle state and action space} that unifies cross-embodiment data format and control problems. For embodiment $e$, we represent the end-effector at time $t$ by a set of $N_e$ particles, $X_t^{(e)} \;=\; \{\, x^{(e)}_{i,t} \in \mathbb{R}^3 \,\}_{i=1}^{N_e}$, the object by $N_o$ particles $X_t^{(o)} \;=\; \{\, x_{i,t} \in \mathbb{R}^3 \,\}_{i=1}^{N_o}$, and thus the world state is represented as \( X_{t} = (X_t^{(e)}, X_t^{(o)}) \). This is a unified particle-based representation applicable to nearly arbitrary end effector (\eg, multi-fingered hands with different DoFs) and objects (\eg, rigid and deformable objects). 

In the particle space, the action can be defined as the end-effector particle displacement field: 
\[
a_t^{P} \;=\; \Delta X_t^{(e)} \;=\; \{\; \delta_{i,t} \in \mathbb{R}^3 \;\}_{i=1}^{N_e},
\]
with $X_{t+1}^{(e)} = X_t^{(e)} + \Delta X_t^{(e)}$. This action information can be computed from passive human-object or robot-object interaction data. We can thus train a world model $f$ to approximate the true transition function $\T$ via supervised learning, which predicts the next state given the action: 
\[
\hat{X}_{t+1} \;=\; \hat f_\theta\!\left(X_t,\, a_t^{P}\right),
\]
and 
\begin{equation}
    \theta^* \;=\; \arg\min_{\theta} \; \mathbb{E}\!\left[\, \mathcal{L}\!\left(\hat f_\theta(X_t, a_t^{P}),\, X_{t+1}\right)\,\right].
    \label{eqn:world-model-training-objective}
\end{equation}
Crucially, this objective requires only state transition data ${X_t, a_t^{P}}_t$, which can be collected from random interactions rather than expert demonstrations. Such data is easier to obtain, and the resulting model can be reused across different task objectives through planning, as described below.

For planning, we obtain particle representations from joint states via forward kinematics (FK). 
Let $\Phi_e:\mathbb{R}^{n_e}\to(\mathbb{R}^3)^{N_e}$ denote the FK mapping for embodiment $e$. 
Given the current and next joint states, $q_t$ and $q_{t+1}=q_t+u_t$, the corresponding particle sets are 
\[
X_t^{(e)} = \Phi_e(q_t), 
\qquad 
X_{t+1}^{(e)} = \Phi_e(q_{t+1}).
\]
The shared particle action is then the displacement field
\[
a_t^{P} = X_{t+1}^{(e)} - X_t^{(e)} \;\in\; (\mathbb{R}^3)^{N_e}.
\]

Planning and learning therefore operate in the embodiment-agnostic state space 
\(
\mathcal{S}^{P}=(\mathbb{R}^3)^{N_e}\times(\mathbb{R}^3)^{N_o}
\)
and action space
\(
\mathcal{A}^{P}=(\mathbb{R}^3)^{N_e}.
\)
This abstraction enables training on data from diverse embodiments and deployment across different hardware without assumptions about the underlying kinematic structure (\eg, degrees of freedom). The only requirement is a forward kinematics model to map joint actions into the particle action space for model inference, which is standard and trivial since robot kinematics are known at deployment. We illustrate the overall framework in \figref{fig:framework}.

\subsection{Perception Module} \label{sec:perception}

The perception module performs state estimation for data collection and deployment. We use a multi-view camera setup following prior work~\cite{robopack, robocook, robocraft, tian2025uniclothdiff}. Cameras are placed at fixed positions around the scene for comprehensive viewpoint coverage.

For human data collection, we reconstruct hand meshes from multi-view images using POEM-v2~\cite{poem-v2} and sample particles with farthest point sampling (FPS). For deformable object perception, we fuse multi-view point clouds, perform Poisson surface reconstruction to obtain a smooth surface~\cite{kazhdan2006poisson}, and apply FPS. For rigid bodies, we estimate object poses using FoundationPose~\cite{foundationposewen2024} and sample surface particles via FPS. Background is excluded from the scene.

During deployment, the robot's state is available from proprioception, and only object perception is required. We apply the same object perception procedure as in data collection.

\subsection{World Model Architecture} \label{sec:world-model-details}
We consider adopting graph neural networks (GNNs) as our world model architecture, as the locality and equivariance are useful inductive biases~\cite{Leslie2020thefoundation} that allow the learned model to generalize to objects and hands with different shapes. We use DPI-Net~\cite{dpi-net}, a GNN that models local particle interactions through message passing and captures global effects via multi-step hierarchical propagation.


Specifically, the graph state at each time step is represented as the tuple $\langle X_t, E_t \rangle$ with $X_t$ as vertices and $E_t$ as edges constructed with a radius graph. For each particle in the graph, $o_{t, i} = \langle x_{i, t}, c_{i, t}^o\rangle$, where $x_{i, t}$ is the particle position $i$ at time $t$, and $c_{i, t}^o$ is the particle's attributes at time t, including the group information (\eg, point belongs to robot or object). In addition, edges between particles are denoted as $e_k = \langle u_k, v_k \rangle$, where $1 \le u_k, v_k \le |O_t|$ are the receiver and sender particle indices respectively. Given the graph, where particles are connected only within a certain radius, we can first use node encoder $f_O^{enc}$ and edge encoder $f_E^{enc}$ to extract node and edge features:
\[
c_{i, t}^o = f_O^{enc}(o_{i,t}), c_{k, t}^e = f_E^{enc}(o_{{u_k},t}, o_{{v_k},t}, d_k^r)
\]
where $d_k^r$ denotes edge's attributes (e.g. length). Then, the features are propagated through edges in multiple steps. Denote $\epsilon_{k,t}^l$ and $h_{i,t}^l$ are propagating influence from edge $k$ and node $i$ at step $l$, respectively. At step 0, initialize $h_{i,t}^0 = 0, i = 1...|O|$. For each step $1\le l \le L$:
\begin{align*}
    \epsilon_{k,t}^l = f_E(c_{k, t}^e, h_{u_k,t}^{l-1}, h_{v_k,t}^{l-1}), k =1 ...|E| \\
    h_{i,t}^l = f_O(c_{i, t}^o, \sum_{k \in \mathcal{N}_i}\epsilon_{k,t}^l, h_{i,t}^{l-1}), i = 1...|O|
\end{align*}
where $\mathcal{N}_i$ is the neighbor index set of particle $i$, $f_O$ denotes the node propagator, and $f_E$ denotes the edge propagator. Then the future state at time $t+1$ is predicted as 
\[
\hat{o}_{i, t+1} = f_O^{dec}(h_{i,t}^L), i = 1...|O|
\]

The particle-based graph network incorporates strong inductive biases. Spatial locality is enforced by restricting message passing to local neighborhoods, analogous to short-range force interactions between particles in physics. Equivariance is achieved through relative coordinates and shared update functions, ensuring invariance to global translations, rotations, and particle permutations. These properties support generalization across embodiments. The model is trained with a supervised objective:
\begin{equation}
\mathcal{L}(O_t, \hat{O}_t) = \ell(O_t, \hat{O}_t),  \label{eqn:loss}
\end{equation}
where $\ell$ denotes the loss function. In simulation, mean squared error (MSE) can be used when temporal point-level correspondence is available, while Chamfer Distance (CD) or Earth Mover’s Distance (EMD) can be applied for unpaired point sets. Thus, the learning objective is broadly applicable.

\subsection{Model-Based Planning} \label{sec:control}



Given learned world models, we use sampling-based model-predictive control to search for optimal trajectories for execution. We devise motion primitives for efficient joint action sampling, inspired by the insight that human hand motions lie in low-dimensional manifolds of the full configuration space~\cite{grasp_taxonomy}. 


For the \pushingtask{} task, we constrain pushing actions to a fixed $x$–$y$ plane. Global translations are sampled as random motion noise in the end-effector frame. Specifically, we first sample a straight-line trajectory in the end-effector frame and then perturb it with Gaussian noise to obtain the final action. The number of fingers making contact with the box is randomly selected.


For the \reshapingtask{} task, we sample joint actions from the following motion primitives, which capture a useful subset of the joint action space:
(i) \fingerspinch{}, involving rotation about the $z$-axis and relative motion between the index finger and thumb;  
(ii) \palmpress{}, characterized by rotation about the $z$-axis and translation along the $z$-axis;  
(iii) \thumbpinch, composed of rotation about the $z$-axis and actuation of thumb-specific degrees of freedom. 

For model-based control, we map sampled joint actions $\{u_t\}_{t=0}^{H-1}$ 
to particles in the shared state space through forward kinematics, rolled out with the learned world model, and evaluated using the cost function. The target is specified as a point cloud following prior work~\cite{ai2025review, robopack, robocook}, and the cost function is defined as a combination of CD and EMD, consistent with the training objective \eqnref{eqn:loss}, \ie, 
\begin{align*}
    \mathcal{J}(X,\mathcal{G}) &= \mathcal{L}_{\mathrm{CD}}(X,\mathcal{G}) + \mathcal{L}_{\mathrm{EMD}}(X,\mathcal{G}) 
\end{align*}
where $X=\{x_i\}_{i=1}^{N}$ denotes the predicted particle point cloud, and $\mathcal{G}=\{g_i\}_{i=1}^{N}$ denotes the target point cloud.

%% file: text/4_experiments.tex
\section{Experiments}

\begin{figure*}[t]
    \centering
    \includegraphics[width=0.9\linewidth]{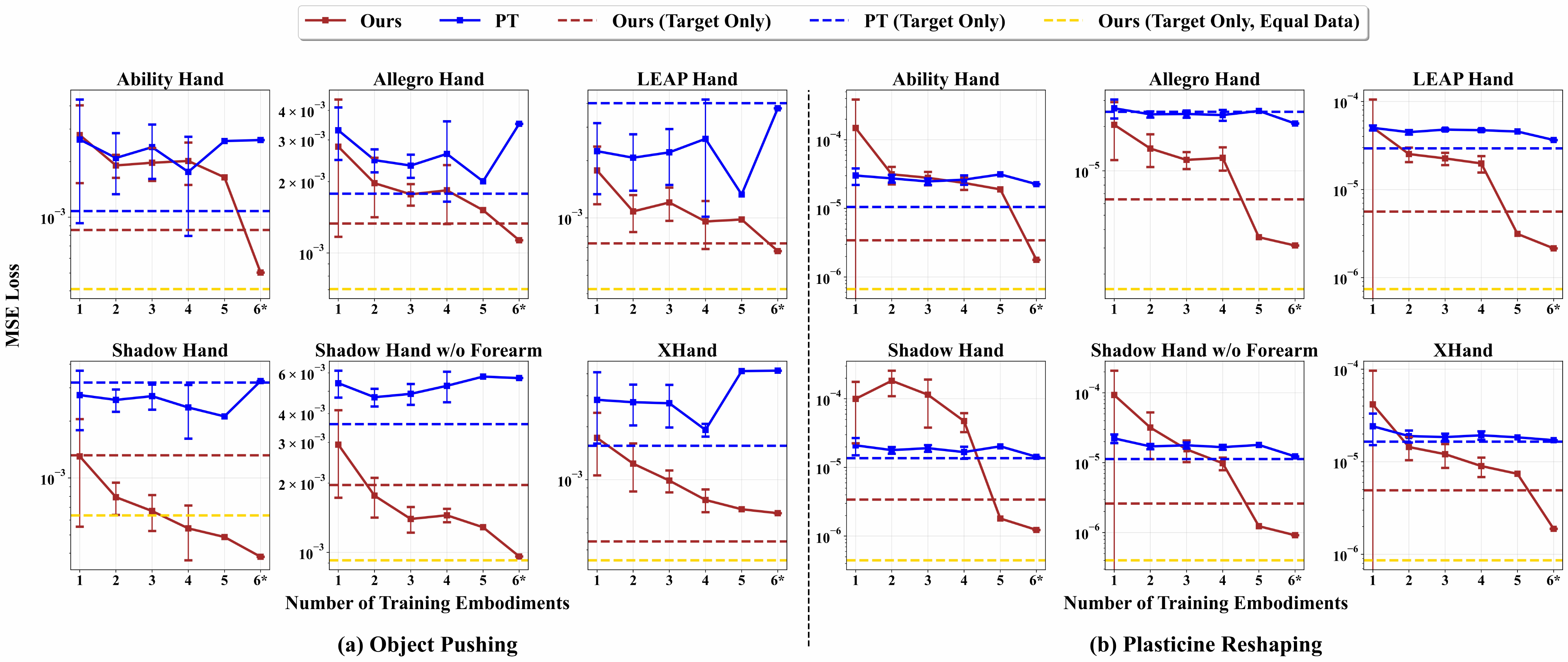} 
    \caption{\textbf{Scaling trends in cross-embodiment world model learning. }
    For each target hand, models are trained on subsets of the remaining hands of varying sizes. 
    All subset combinations at a given size are enumerated (\eg, $\binom{5}{2}$ for size 2), and the mean performance with 95\% confidence intervals is reported. 
    Dashed lines indicate models directly trained on the target embodiment. 
    ``Equal data'' refers to training directly on the target embodiment using the same number of samples as the total data aggregated across all source hands.
    Across both tasks, our model exhibits a embodiment scaling trend: increasing the number of training embodiments consistently lowers prediction error, and with five source embodiments (zero-shot) it often matches or surpasses target-only training while Point Transformer (PT)~\cite{zhao2021point} shows weaker and less consistent gains, potentially due to the lack of inductive bias. 
    }
    \label{fig:scaling_generalization}
    \vspace{-10pt}
\end{figure*}

\begin{figure}[t]
    \centering
    \includegraphics[width=\columnwidth]{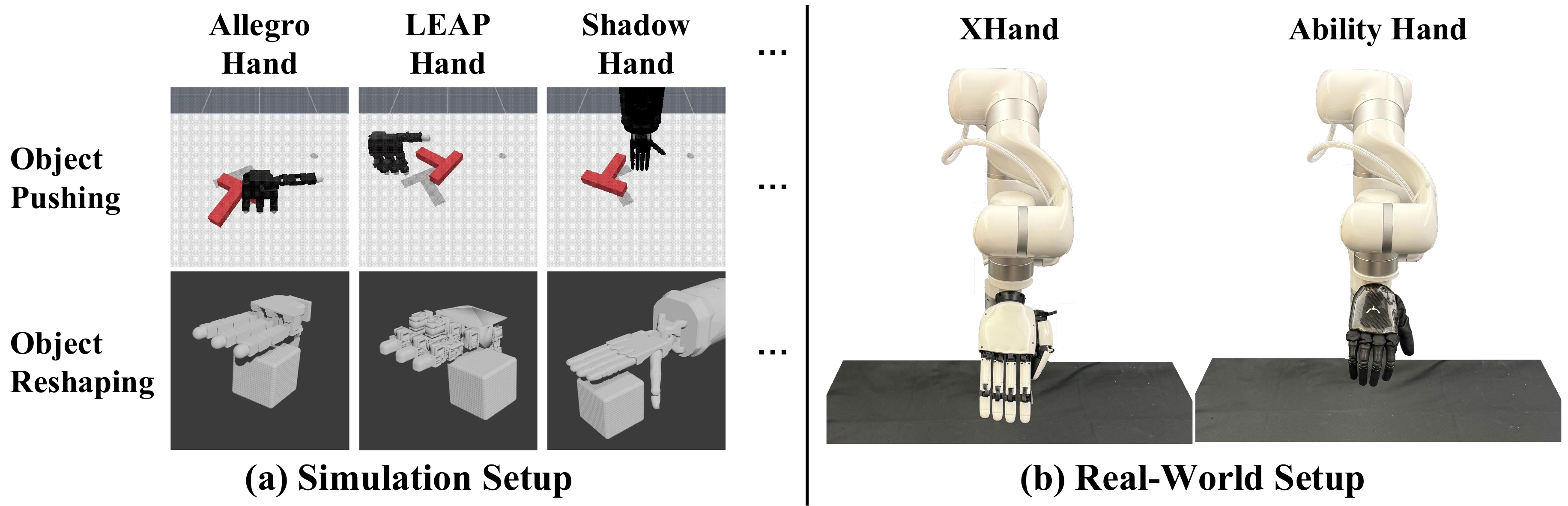} \vspace{-5pt} 
    \caption{\textbf{Cross-embodiment setups in simulation and the real world.} 
            We have multiple robotic hands in simulation for collecting random interaction data, and two real hardware mounted on a UFACTORY XArm 7 for system deployment. }
    \label{fig:setup} \vspace{-13pt}
\end{figure} 

In this section, we study the following questions: 

\begin{enumerate}[nosep]
    \item[\textbf{Q1.}] Does cross-embodiment training of the world model improve generalization on unseen embodiments? 
    \item[\textbf{Q2.}] What is the co-training recipe to leverage simulation and real-world data? 
    \item[\textbf{Q3.}] Does the learned dynamics model enable effective planning for dexterous manipulation?
\end{enumerate}

Our study proceeds in three stages. First, we investigate cross-embodiment scaling entirely in simulation (\qone{}), which provides a clean and controlled environment for studying scaling behaviors. Second, to bridge the sim-to-real gap and capture realistic contact dynamics, we incorporate real human data, and study the training recipe to best leverage the simulation robot data and real human data (\qtwo{}). Finally, we evaluate the trained world models on real robot hardware to assess their quality at the system level (\qthree{}).

\subsection{Experimental Setup}

\textbf{Task setup. } 
We consider two representative dexterous manipulation tasks: non-prehensile rigid object pushing~\cite{robopack, chi2024diffusion} and deformable object reshaping~\cite{robopack, robocook, robocraft}. In \pushingtask{}, the goal is to reorient a box to a target orientation. In \reshapingtask{}, the goal is to mold plasticine into a target shape. The targets are specified as point cloud observations, following~\cite{robocook, robopack}. Both tasks require reasoning about object dynamics and precise contact control. 

\textbf{Model implementation.}
For \reshapingtask{}, we represent the object using 300 particles and the hand using 200 particles. We construct radius graphs with an inner radius of 0.025 m and an outer radius of 0.04 m~\cite{robocraft}. For \pushingtask{}, we represent the object using 100 particles and the hand using 50 particles, with both radii set to 0.04 m. We use a higher particle density for \reshapingtask{} to capture finer-grained local contact patterns.

At each MPC iteration, we sample 500 candidate action sequences with a planning horizon of 4 steps and perform 10 optimization iterations. We execute the first 2 action steps before replanning. On a workstation equipped with an RTX 4090 GPU, each planning update takes approximately 60 s.

\textbf{Baseline.}
We adapt the Point Transformer (PT)~\cite{zhao2021point} as an alternative architecture for modeling particle dynamics. Unlike DPI-Net, which uses structured message passing, PT models interactions among particles using an attention mechanism. PT and DPI-Net share the same input and output formats and are trained on the same data using the same loss functions to ensure a fair comparison. Because we collect only random interaction data, we exclude approaches that require expert demonstrations~\cite{chi2024diffusion, intelligence2026pi, foar}.

\textbf{Simulation setup. }  
We simulate six dexterous hands representative of commonly used multi-fingered designs: Ability Hand (6-DoF), Allegro Hand (16-DoF), XHand (12-DoF), Leap Hand (16-DoF)~\cite{leaphand}, Shadow Hand (24-DoF), and a URDF variant of the Shadow Hand without its forearm (24-DoF).  
For the rigid-body task (\pushingtask{}), we use SAPIEN~\cite{Xiang_2020_SAPIEN} for data collection.  
For deformable object manipulation, we use the Rewarped simulation platform~\cite{rewarped}, a differentiable multiphysics simulator.  
We collect 100 trajectories per task for each robots, where the robots perform random actions in the predefined action space.

\textbf{Real-world setup. }  
Our hardware platform consists of a 7-DoF XArm robot equipped with an Ability Hand and an XHand. Four Intel RealSense cameras provide multi-view perception, following prior work~\cite{robocook, robocraft, robopack}. The system is controlled via a workstation with an NVIDIA RTX 4090 GPU. For human demonstration data, we collect 30 minutes of demonstrations for \thumbpinch{}, \fingerspinch{}, and \palmpress{} each. The simulation and real-world hardware setup are illustrated in \figref{fig:setup}.

\subsection{Evaluating Cross-Embodiment World Model Learning}

\begin{figure*}[t]
    \centering
    \includegraphics[width=0.8\textwidth]{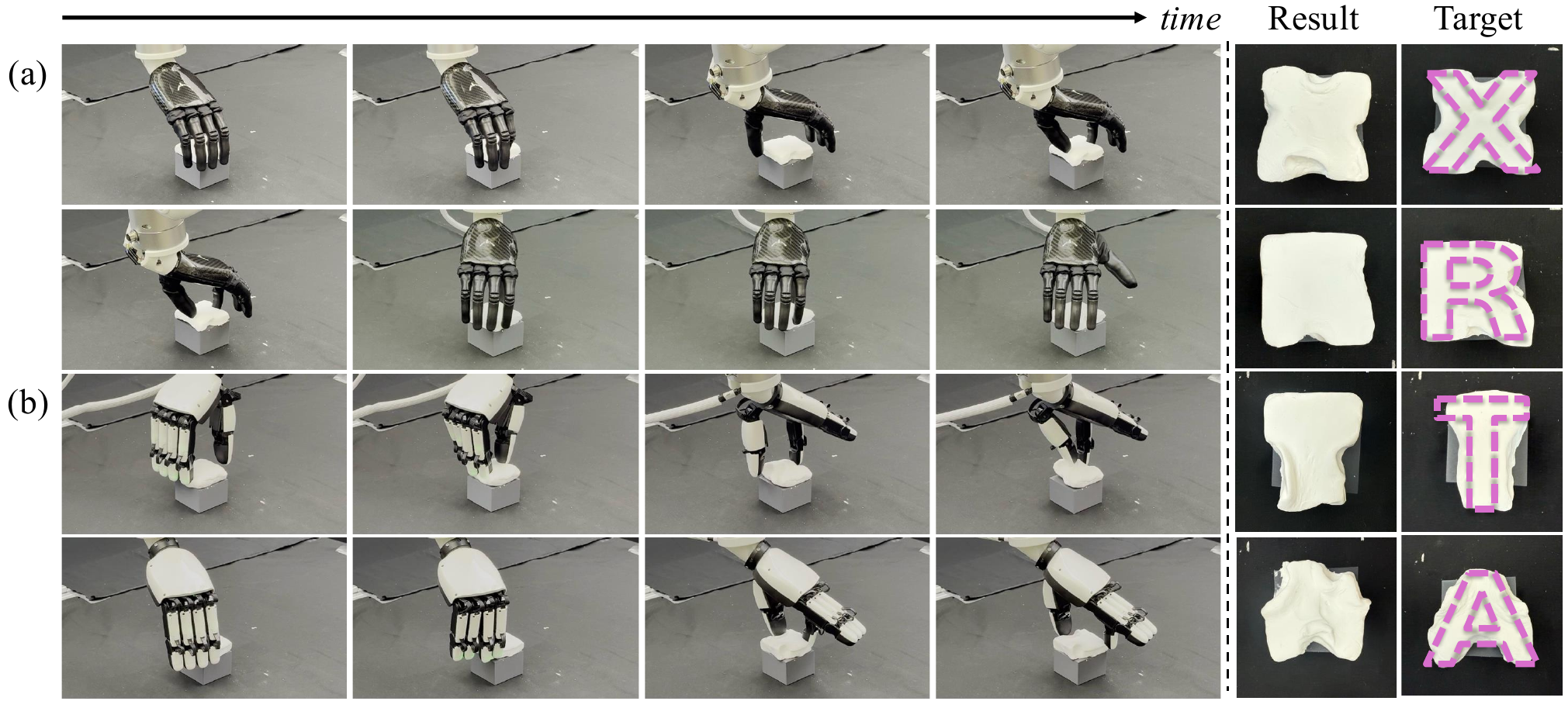}
    \caption{\textbf{Qualitative results of cross-embodiment deployment}. (a) Ability Hand (6-DoF) and (b) XHand (12-DoF) utilize the same particle‑space dynamics model learned from human demonstration. For each trial, the hand successfully reshapes the deformable clay toward the target shape using a combination of \fingerspinch{}, \palmpress{}, and \thumbpinch{} skills.}
    \label{fig:qualitative} \vspace{-15pt}
\end{figure*}

\begin{figure}[t]
    \centering
    \includegraphics[width=\columnwidth]{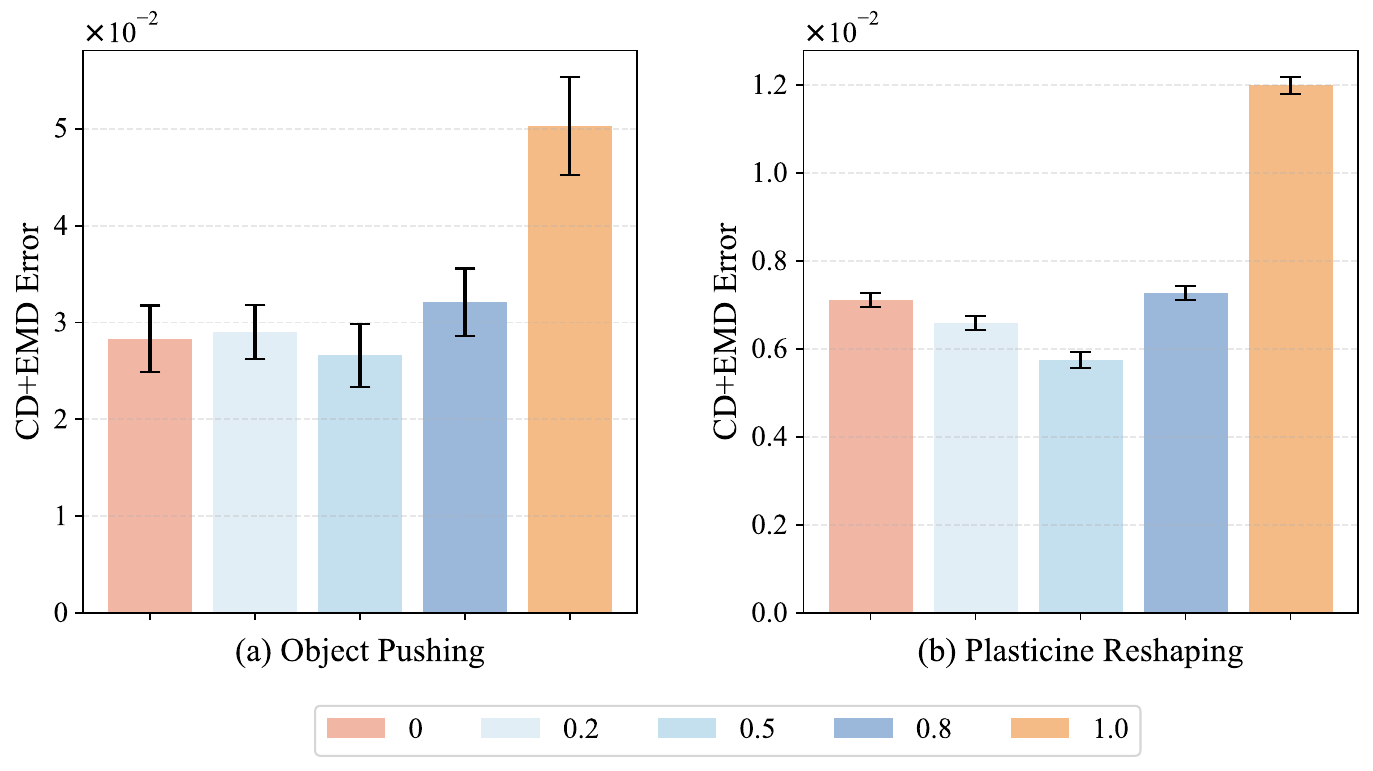}
    \caption{\textbf{Evaluating training recipes for bridging simulation and real.}
    We compare co-training with different mixtures of simulation and real-world data. Legend values indicate the amount of simulation data relative to a fixed quantity of real human data. The y-axis shows prediction error on held-out human interactions, with error bars denoting 95\% confidence intervals. Here, “CD+EMD” denotes an equally weighted sum (CD+EMD, 1:1). We adopt CD/EMD since the real-world data lacks temporal point-level correspondence.
    }
    \label{fig:recipe} 
    \vspace{-13pt}
\end{figure}

We systematically evaluate how the number of training embodiments influences generalization to unseen embodiments. For each target hand, we hold it out and train on $x$ other hands, enumerating all $\binom{N}{x}$ subsets from $N=6$ total hands. The mean squared error (MSE) on the unseen hand serves as the generalization metric. We report results for both our method and a Point Transformer (PT)~\cite{zhao2021point} baseline trained under the same subset protocol and data budget. In addition, the case $x=6$ corresponds to training on all hands, including the target, and provides a reference for the upper bound of cross-embodiment learning in the current data regime. Results are shown in \figref{fig:scaling_generalization}.

\textbf{Key observations.}  We make the following observations:
\begin{itemize}
    \item \emph{Embodiment scaling law~\cite{ai2025towards}:} Prediction error decreases as more embodiments are included, and variance across subsets shrinks, indicating more stable models with broader embodiment diversity for our GNN. In contrast, PT~\cite{zhao2021point} shows weaker and less consistent improvements as $x$ increases, suggesting that the inductive bias in GNN contributes to the cross-embodiment transfer.
    \item \emph{Zero-shot strength at $x{=}5$:} With five training embodiments (no target data), the performance of GNN often approaches or surpasses training directly on the target hand. This shows that diverse cross-embodiment data can substitute for target-specific data when deploying to a new hand. This opens up the possibility of building cross-embodiment generalist world models that can broadly zero-shot transfer to novel ones via large-scale cross-embodiment training.
    \item \emph{Benefit of co-training at $x{=}6^\ast$:} Even when target data is available, adding the other embodiments yields further gains over target-only training. Our proposed state and action representations unify data from heterogeneous embodiments and make such co-training possible; PT~\cite{zhao2021point} benefits less from co-training, consistent with its higher sensitivity to which training subset is used.
\end{itemize}

\begin{table*}[t]
\centering
\setlength{\tabcolsep}{6pt}
\begin{tabular*}{0.9\textwidth}{@{\extracolsep{\fill}} l l l c c c c c @{}}
\toprule
Hand Type & Method & Metric & Letter \texttt{X} & Letter \texttt{R} & Letter \texttt{T} & Letter \texttt{A} & Aggregated \\
\midrule

\multirow{6}{*}{Ability Hand}
  & \multirow{3}{*}{\textbf{Co-train}}
    & CD $\times 10^{-3}$ $\downarrow$
      & 6.85 $\pm$ 0.21
      & 6.92 $\pm$ 0.10
      & 6.88 $\pm$ 0.19
      & 7.14 $\pm$ 0.14
      & \textbf{6.95 $\pm$ 0.10} \\
  & &
    EMD $\times 10^{-3}$ $\downarrow$
      & 4.76 $\pm$ 0.17
      & 4.90 $\pm$ 0.35
      & 4.94 $\pm$ 0.27
      & 5.10 $\pm$ 0.18
      & \textbf{4.92 $\pm$ 0.13} \\
  & &
    Success Rate $\uparrow$
      & 5 / 5
      & 4 / 5
      & 5 / 5
      & 4 / 5
      & \textbf{18 / 20} \\
\cmidrule(lr){2-8}

  & \multirow{3}{*}{Human-only}
    & CD $\times 10^{-3}$ $\downarrow$
      & 7.21 $\pm$ 0.55
      & 7.20 $\pm$ 0.15
      & 6.94 $\pm$ 0.14
      & 7.26 $\pm$ 0.14
      & 7.15 $\pm$ 0.16 \\
  & &
    EMD $\times 10^{-3}$ $\downarrow$
      & 5.07 $\pm$ 0.67
      & 5.38 $\pm$ 0.39
      & 5.08 $\pm$ 0.25
      & 5.40 $\pm$ 0.36
      & 5.23 $\pm$ 0.23 \\
  & &
    Success Rate $\uparrow$
      & 3 / 5
      & 2 / 5
      & 4 / 5
      & 1 / 5
      & 10 / 20 \\
\midrule

\multirow{6}{*}{XHand}
  & \multirow{3}{*}{\textbf{Co-train}}
    & CD $\times 10^{-3}$ $\downarrow$
      & 6.65 $\pm$ 0.26
      & 6.70 $\pm$ 0.11
      & 6.99 $\pm$ 0.17
      & 7.05 $\pm$ 0.20
      & \textbf{6.85 $\pm$ 0.12} \\
  & &
    EMD $\times 10^{-3}$ $\downarrow$
      & 4.52 $\pm$ 0.29
      & 4.65 $\pm$ 0.17
      & 4.90 $\pm$ 0.21
      & 5.07 $\pm$ 0.23
      & \textbf{4.78 $\pm$ 0.14} \\
  & &
    Success Rate $\uparrow$
      & 5 / 5
      & 5 / 5
      & 4 / 5
      & 3 / 5
      & \textbf{17 / 20} \\
\cmidrule(lr){2-8}

  & \multirow{3}{*}{Human-only}
    & CD $\times 10^{-3}$ $\downarrow$
      & 6.82 $\pm$ 0.30
      & 7.22 $\pm$ 0.27
      & 7.32 $\pm$ 0.31
      & 7.53 $\pm$ 0.32
      & 7.22 $\pm$ 0.18 \\
  & &
    EMD $\times 10^{-3}$ $\downarrow$
      & 4.66 $\pm$ 0.44
      & 5.26 $\pm$ 0.31
      & 5.37 $\pm$ 0.29
      & 5.43 $\pm$ 0.33
      & 5.18 $\pm$ 0.21 \\
  & &
    Success Rate $\uparrow$
      & 4 / 5
      & 2 / 5
      & 2 / 5
      & 1 / 5
      & 9 / 20 \\
\bottomrule
\end{tabular*}
\caption{\textbf{Quantitative results of cross-embodiment deployment on Plasticine Reshaping task.} Real-world performance comparison of co-training (human + 6 simulated robot hands) vs.\ training on only human data, evaluated on Ability Hand and XHand. Columns correspond to target letter \texttt{X}/\texttt{R}/\texttt{T}/\texttt{A} settings. Reported values are mean $\pm$ 95\% confidence interval. }
\label{tab:system}\vspace{-13pt}
\end{table*}

\textbf{Task-specific differences.}  
Errors are generally lower for deformable reshaping, as deformations are spatially localized, whereas rigid-body rotations move particles over a much larger scale. At the same time, the scaling effect is more pronounced in deformable manipulation. We hypothesize this is because deformable tasks involve larger contact surfaces, making end-effector geometry more influential. Exposure to diverse embodiments therefore provides richer coverage of contact geometries and interaction patterns, which aids generalization. By contrast, rigid pushing often depends on a small number of contact points, where cross-embodiment differences are less critical. These results suggest that our approach is particularly beneficial for tasks with complex contact dynamics, as in \reshapingtask{}.

\textbf{Embodiment-specific trends.}
Certain hands (\eg, Shadow Hand, Leap Hand) show sharp improvements when scaling from 4 to 5 training embodiments, whereas smaller hands (\eg, Ability Hand) achieve competitive performance earlier. We hypothesize that this effect is linked to graph density in the particle–graph representation used by our GNN-based world model. Smaller hands have fewer degrees of freedom but a more compact geometry, which results in denser particle connections under the radius-graph construction. This denser connectivity provides richer local message passing and allows the GNN to propagate interaction information more effectively, even when trained on fewer embodiments. By contrast, larger hands span a larger spatial extent, yielding sparser graphs where local neighborhoods capture fewer interactions. In such cases, broader embodiment diversity is needed to expose the model to sufficient variations in contact patterns and fill in the missing structural information. This suggests that the scaling benefits of adding more training embodiments are not uniform across morphologies, but depend on the intrinsic graph density of each hand’s particle representation. We believe developing model architectures that are less sensitive to graph densities is an interesting future direction. 

\subsection{Co-Training Recipe}

Having established positive embodiment scaling in simulation, we next study how to leverage simulation data for real-world learning. 
Simulation offers uniform sensing and abundant interactions, but models trained purely in simulation can overfit to simulator-specific artifacts such as contact or material mismatches. 
Conversely, real-world human data avoids the reality gap but introduces an embodiment gap relative to robot hands. 
We hypothesize that co-training on both domains may combine their complementary strengths, when the signals from each are balanced appropriately.  

We train models with different mixtures of simulation and real-world data, and evaluate them on held-out real human data (\figref{fig:recipe}). 
Simulation-only training yields the highest prediction error, highlighting the sim-to-real gap. 
Human-only training provides a stronger baseline, and mixing simulation with human data further reduces error when the ratio is well balanced. 
Notably, a 1:1 ratio performs best across tasks, suggesting that simulation data can act as a useful regularizer for human data rather than a substitute. 


\subsection{Evaluating Model-Based Control}

For real-world deployment, we focus on the more challenging task \reshapingtask{}, which has complex contact dynamics. We compare two models: one trained on human data only and the best-performing co-training model. Each model is evaluated across four target shapes (``\texttt{X}'', ``\texttt{R}'', ``\texttt{T}'', ``\texttt{A}''), with five trials per shape, for a total of 20 runs. We report CD, EMD, and success rates, where success is defined as achieving an error lower than 0.0125 for CD $+$ EMD loss.
Quantitative results are reported in \tabref{tab:system}. 
The human-only model achieves zero-shot transfer to novel robot hands by leveraging the unified state and action space, but its performance is lower than that of the co-training model due to the lack of embodiment diversity in the training data.  
On Ability Hand, co-training achieves $18/20$ successes, while the human-only model reaches only $10/20$ successes. The improvement is especially visible on letters \texttt{A} and \texttt{R} where human-only struggles. On XHand, co-training similarly attains $17/20$ successes, compared to $9/20$ for human-only. Co-training is robust across targets (notably \texttt{R} and \texttt{X} at $5/5$ each), whereas human-only is much more target-dependent.

Qualitative results of the co-training model are shown in \figref{fig:qualitative}. Both the \textit{(a)} Ability Hand and \textit{(b)} XHand successfully reshape clay into target letters by composing the three predefined motion skills, \thumbpinch{}, \fingerspinch{}, and \palmpress{}, to carve, spread, and compress. Despite their kinematic differences, the same particle-based dynamics model enables model-predictive planning on both hands without fine-tuning, demonstrating effective cross-embodiment deployment.

%% file: text/6_conclusions.tex
\section{Conclusion \& Discussion}

This work positions world models as a shared interface for learning and control across embodiments. By representing heterogeneous hands and their actions in a unified particle space, we train a single dynamics model from simulated robot interactions and real human interactions and deploy it for model-predictive control on distinct robotic hands. Our experiments show that prediction and control improve as the diversity of training embodiments increases, and that appropriately combining simulation and real-world data outperforms either source alone. Together, these results suggest that the transferable structure across embodiments lies not in their joint spaces, but in the physical interactions they induce in the world. World models that capture this shared structure therefore offer a promising path toward generalist systems that learn from heterogeneous embodiments and control new ones without embodiment-specific policy training.